\documentclass[letterpaper, 10 pt, conference]{ieeeconf}  

\usepackage{amsmath} 
\usepackage{amssymb}  
\usepackage{caption}
\usepackage{multirow}
\usepackage{pifont}
\usepackage{array}
\usepackage{subfigure}
\usepackage{graphicx}
\usepackage{balance}
\usepackage{times}
\usepackage{dsfont}
\usepackage{multicol}
\usepackage[per-mode = symbol]{siunitx}
\usepackage{xspace}
\usepackage{xurl}
\usepackage{capt-of}
 \usepackage[noadjust]{cite}
\usepackage{wrapfig,lipsum,booktabs}
\usepackage{booktabs}
\usepackage[dvipsnames]{xcolor}
\usepackage{hyperref}
\usepackage{threeparttable}

\definecolor{mydarkblue}{rgb}{0,0.08,0.45}
\definecolor{mydarkgreen}{RGB}{0, 139, 69}
\definecolor{mygreen2}{RGB}{0 205 0}
\definecolor{mybrown}{RGB}{139 69 19}
\definecolor{myred}{RGB}{188, 0, 0}
\definecolor{myblue}{RGB}{0, 0, 198}
\definecolor{ourcolor}{RGB}{118,10,0}
\definecolor{ourglowcolor}{RGB}{230,190,0}
\definecolor{baselinecolor}{RGB}{225,225,0} 
\definecolor{baselineglowcolor}{RGB}{200,0,255} 
\definecolor{boxblue}{RGB}{79,173,234}
\definecolor{boxgreen}{RGB}{159,206,99}

\title{\LARGE \bf
SEEC: Stable End-Effector Control with Model-Enhanced 

Residual Learning for Humanoid Loco-Manipulation
}

\author{Jaehwi Jang$^{*}$\textsuperscript{1}, Zhuoheng Wang$^{*}$\textsuperscript{1,2}, Ziyi Zhou\textsuperscript{1}, Feiyang Wu\textsuperscript{1}, and Ye Zhao\textsuperscript{1}\\
\textsuperscript{*}Equal Contribution \quad \textsuperscript{1}Georgia Institute of Technology \quad \textsuperscript{2}Tsinghua University\\
}
\thanks{* Equal Contribution. Project Co-lead.}
\thanks{$^{1}$ Tsinghua University, 100084, China
        }%
\thanks{$^{2}$ Stanford University, CA
        }%

\begin{document}

\twocolumn[{%
\renewcommand\twocolumn[1][]{#1}%
\maketitle
\begin{center}
    \vspace{-0.1in}
    \centering
    \captionsetup{type=figure}
    \includegraphics[width=\linewidth]{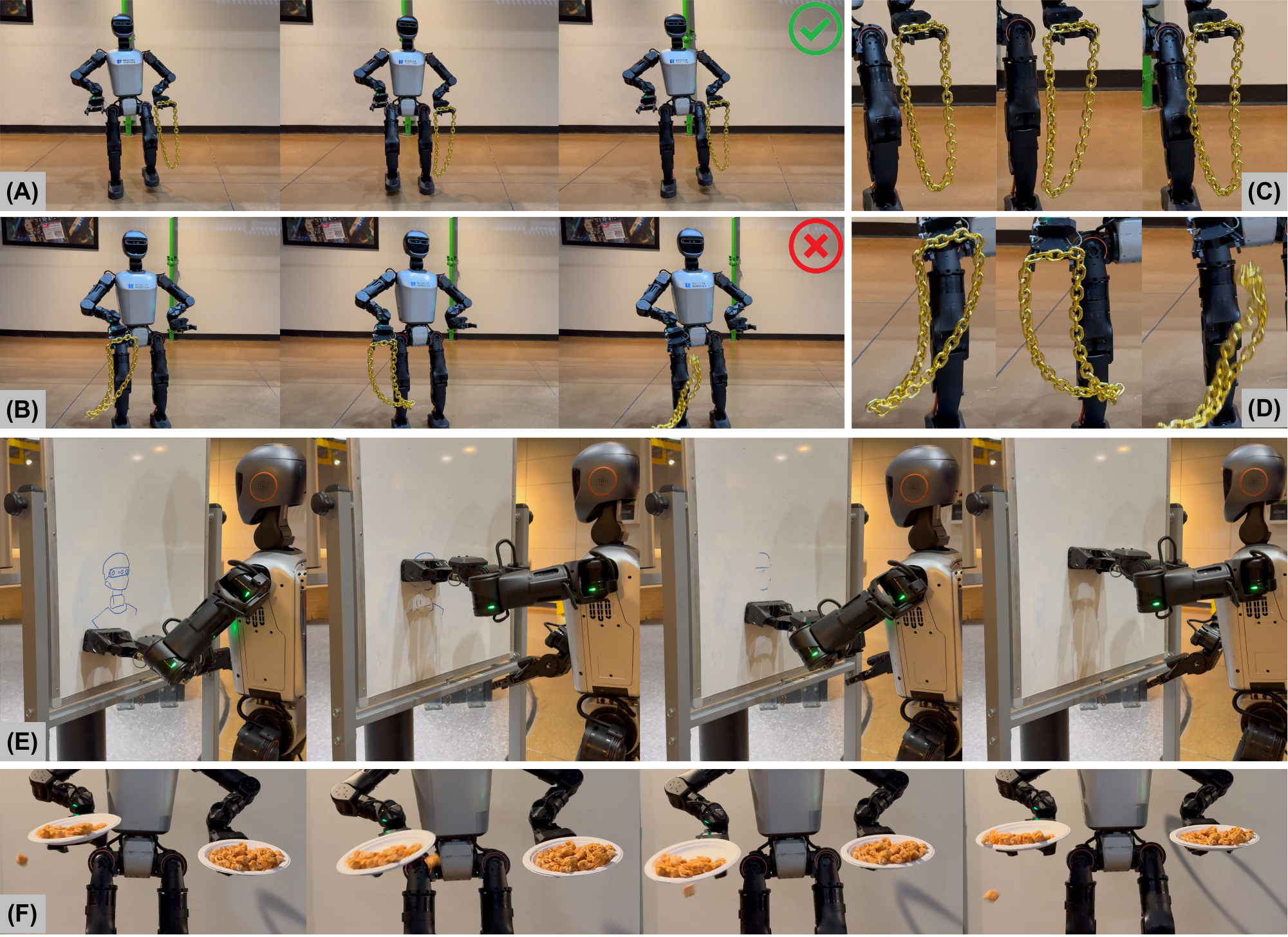}
    \captionof{figure}{Our SEEC framework enables a Booster T1 humanoid robot to perform stable loco-manipulation tasks while dynamic locomotion. Demonstrated skills include (A-D) holding a flexible chain while walking, (E) wiping a whiteboard surface with teleoperation, and (F) carrying a plate of snacks while walking. In (A-D), our SEEC framework (A, C) enables the robot to firmly hold the chain and suppress oscillatory dynamics, whereas the IK baseline (B, D) induces large oscillations that eventually cause the chain to drop. In (F), our SEEC framework (left arm) successfully kept the snacks in the plate, whereas the IK baseline (right arm) failed and dropped the snacks. }
    \label{fig:teaser}
\end{center}%
}]

\thispagestyle{empty}
\pagestyle{empty}

\begin{abstract}
Arm end-effector stabilization is essential for humanoid loco-manipulation tasks, yet it remains challenging due to the high degrees of freedom and inherent dynamic instability of bipedal robot structures. Previous model-based controllers achieve precise end-effector control but rely on precise dynamics modeling and estimation, which often struggle to capture real-world factors (e.g., friction and backlash) and thus degrade in practice. On the other hand, learning-based methods can better mitigate these factors via exploration and domain randomization, and have shown potential in real-world use. However, they often overfit to training conditions, requiring retraining with the entire body, and still struggle to adapt to unseen scenarios. 
To address these challenges, we propose a novel stable end-effector control (SEEC) framework with model-enhanced residual learning that learns to achieve precise and robust end-effector compensation for lower-body induced disturbances through model-guided reinforcement learning (RL) with a perturbation generator. This design allows the upper-body policy to achieve accurate end-effector stabilization as well as adapt to unseen locomotion controllers with no additional training.
We validate our framework in different simulators and transfer trained policies to the Booster T1 humanoid robot. Experiments demonstrate that our method consistently outperforms baselines and robustly handles diverse and demanding loco-manipulation tasks.
\end{abstract}

\section{INTRODUCTION}

Humanoid robots promise seamless integration into human environments, where they must walk and manipulate simultaneously. From carrying objects while moving to performing collaborative tasks \cite{gu2025humanoid, he2024omnih2o, su2025hitter}, this capability is fundamental for practical humanoid deployment (see the tasks shown in Fig.~\ref{fig:teaser}). 
Yet, achieving stable and precise control of the arm end-effector during dynamic locomotion remains an open challenge.
Even modest base movements could induce large accelerations at the arm end-effector, causing tracking errors, destabilizing contact forces, and ultimately limiting the utility of humanoids in real-world settings.

Recently, learning-based approaches \cite{fu2023deep, liu2024visual, portela2024whole, li2025softa} have sought to achieve humanoid loco-manipulation by training end-to-end reinforcement learning (RL) policies. 
While effective at capturing nonlinearities and handling uncertainty,
these policies often rely on imitating joint or task reference trajectories  \cite{liu2024opt2skill, cheng2024expressive, he2024hover}, and struggle to ensure accurate end-effector stabilization.
For example, in HOVER \cite{he2024hover}, uncontrolled hand motions emerge as a byproduct of locomotion.
Prior work \cite{li2025softa} has attempted to stabilize end-effector control by directly penalizing its acceleration, but this approach heavily relies on policy optimization to “discover” the right compensation strategy.
Additionally, the learned behavior degenerates into static hand-holding motions, limiting general applicability.
Moreover, when tasks require reactive whole-body coordination, instability in end-effector control is exacerbated by sudden locomotion disturbances. Conventional RL training, as in \cite{li2025softa}, fails to provide robustness under such out-of-distribution (OOD) scenarios.

Inspired by model-based approaches \cite{osman2020end, minniti2019whole, wang2024hierarchical, woolfrey2021predictive, ma2022combining}, which achieve precise stabilization through dynamics modeling and online estimation, we introduce SEEC: \textbf{S}table \textbf{E}nd-\textbf{E}ffector \textbf{C}ontrol, a model-enhanced RL framework for humanoid loco-manipulation.
SEEC leverages model-based expertise to provide analytic acceleration compensation signals during training. Instead of relying on naive penalization, the compensation torque from the model-based formulation is distilled into the RL policy, addressing instability in a more principled manner.

Furthermore, unlike prior works that jointly train manipulation and locomotion policies, we introduce a \textit{perturbation generation strategy} that exposes the upper-body policy to a wide spectrum of locomotion-induced disturbances. By modeling these disturbances as base movement patterns, the upper-body controller learns to maintain stable arm end-effector control independent of any specific locomotion policy.
This modular design not only improves robustness across diverse walking patterns, allowing seamless transfer across different walking patterns, including previously unseen locomotion controllers, but also facilitates integration into complex loco-manipulation tasks that demand coherent whole-body coordination.

Our core contributions can be summarized as follows.
\begin{itemize}
    \item We propose a model-enhanced residual learning framework that integrates model-based expertise with learning-based adaptability, achieving precise acceleration compensation while effectively addressing model inaccuracies and parameter uncertainties.
    \item We introduce a \emph{base-movement data generation and perturbation generation strategy} that exposes the policy to a broad spectrum of locomotion-relevant disturbances during training. This enables the controller to acquire robust compensation behaviors that \emph{can transfer to unseen locomotion controllers and gaits} without requiring joint re-training.
    \item We demonstrate the first deployment of such a hybrid framework on a full humanoid Booster T1, validating it both in simulation and on the real hardware via zero-shot transfer. The system achieves more stable and precise end-effector control across a variety of loco-manipulation tasks, compared to the baselines.
\end{itemize}

\section{RELATED WORK}
Traditional works on whole-body controllers for legged and mobile manipulators rely on model-based methods, which often employ numerical optimization to achieve precise control \cite{wensing2023optimization, sleiman2023versatile, kuindersma2016optimization, bellicoso2019alma}. 
Although effective, these approaches depend on accurate dynamics modeling and contact scheduling, which are difficult to maintain in complex or unstructured environments. 
In contrast, learning-based methods have rapidly advanced humanoid whole-body control, driven by reinforcement learning (RL) and imitation learning (IL) \cite{gu2025humanoid, he2024omnih2o, fu2023deep, portela2024whole, cheng2024expressive, ji2024exbody2, zhuang2024humanoid, wang2025dribble}.
These frameworks produce expressive and robust behaviors, but achieving accurate and stable end-effector control remains a fundamental challenge. The difficulty arises because locomotion-induced disturbances rapidly amplify tracking errors, especially during agile maneuvers or in dynamic environments.
SoFTA \cite{li2025softa} attempts to stabilize the end-effector by penalizing its acceleration in the reward function. However, this approach often degenerates into static hand-holding behaviors and does not generalize to diverse loco-manipulation tasks.
By contrast, we show that compensating for the torque needed to account for locomotion-induced disturbances enables effective stabilization across diverse walking motions, including motions produced by controllers not seen during training.

Effective control of manipulators or arms on mobile/legged robots is crucial to achieve expressive motion and thus solve complex tasks. 
To manage complexity, many recent works adopt a decoupled architecture, splitting into upper-body and lower-body modules \cite{ma2022combining, fu2023deep, cheng2024expressive, pan2025roboduet, li2025softa}.
Ma et al. \cite{ma2022combining} model the influence of the manipulator on locomotion as a disturbance, training the locomotion controller to compensate. 
While this improves gait stability, it sidesteps the harder problem of stabilizing the arm under dynamic base motion.
Moreover, existing frameworks \cite{li2025softa} jointly train locomotion and manipulation policies, coupling them tightly.
This prevents modular reuse and limits robustness: when deployed with different locomotion controllers, or in the face of the inevitable dynamically changing real-world environments, the upper-body policy cannot adapt to unseen disturbances and often fails.
Our framework departs from this paradigm. 
We explicitly model lower-to-upper body coupling and introduce a \textit{perturbation generation strategy}, which independently trains the upper-body controller to compensate for a wide range of locomotion-induced disturbances. 
This design enables stable and precise end-effector control that handles unseen perturbations in the real world and even maintains performance across locomotion controllers.

Recent works have explored how to combine MPC and RL controllers \cite{gu2025humanoid}. 
One line of remarkable research uses a model-based controller or trajectory optimizer to supervise learning, where the RL policies imitate expert actions \cite{liu2024opt2skill, kang2023rl+, youm2023imitating, jung2025preci}.
Another line blends outputs from RL and MPC, using MPC for constraint satisfaction and RL for adaptability \cite{cheng2025rambo, bang2024rl}. These approaches improve sample efficiency and stability, but are generally evaluated on flat terrain or simplified tasks and rarely address the unique challenge of maintaining arm end-effector stability during dynamic loco-manipulation.
In this work, we adopt a residual policy learning approach \cite{silver2018residual, cheng2025rambo}, but tailored for humanoid loco-manipulation, achieving robust end-effector stabilization.
\begin{figure*}[t]
    \centering
    \includegraphics[width=1.0\linewidth]{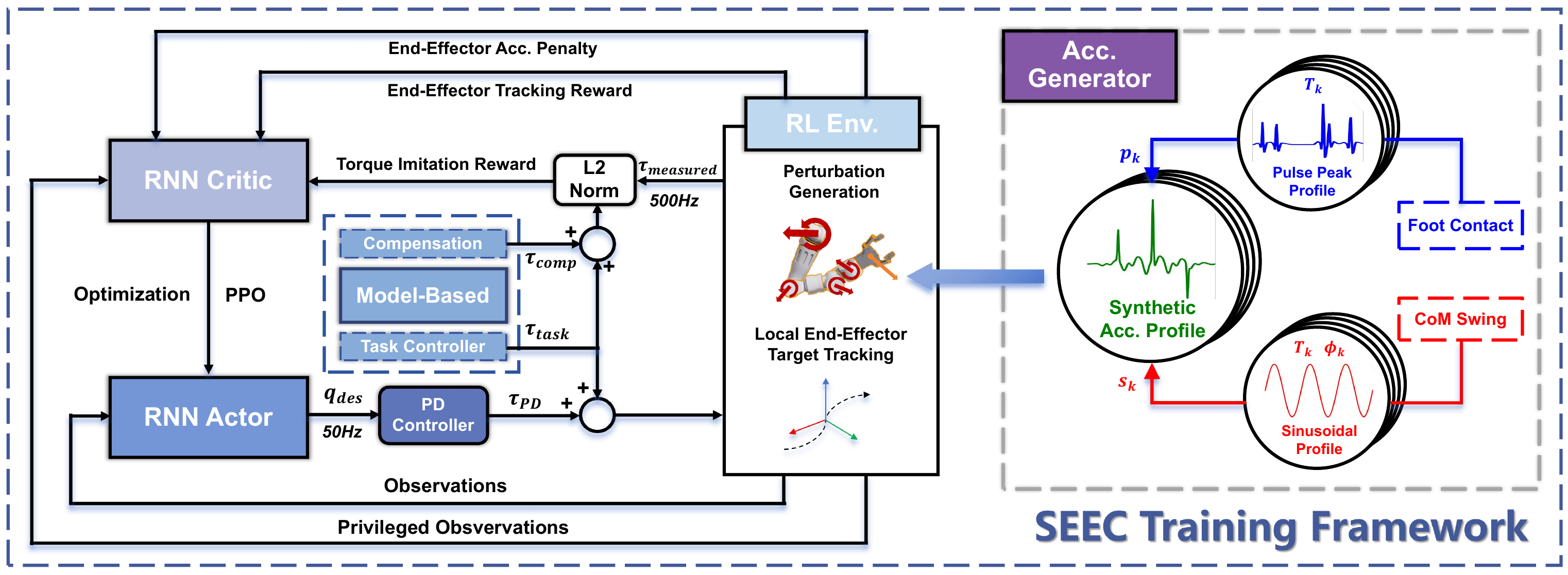}
    \caption{\textbf{System framework overview of {SEEC}.} Our SEEC framework decouples the humanoid loco-manipulation controller into upper-body and lower-body controllers. The figure describes our core upper-body reinforcement learning (RL) module, which trains a residual policy that compensates lower-body induced disturbances. We leverage model-based acceleration compensation signals to guide RL training, ensuring more principled end-effector stability than naive penalization. Additionally, we generate base acceleration profiles to simulate external perturbations to promote robustness to unseen locomotion controllers. For the deployment, we transfer trained upper-body and lower-body policies to the robot without additional joint training.
    }
    \label{fig:framework}
\end{figure*}
\section{METHOD}

We formulate our control problem as the coordination of two controllers:  
(i) a \emph{lower-body controller} responsible for locomotion, and  
(ii) an \emph{upper–body controller} responsible for manipulation tasks.  
Both policies are trained in IsaacLab \cite{mittal2023orbit} and modeled as Markov Decision Processes (MDPs). At time $t$, the agent (each policy) receives observation $o_t$, and then samples an action $a_t \sim \pi(\cdot|o_t)$ according to policy $\pi$ and transitions to a new observation $o_{t+1}$ while receiving a reward $r(o_t, a_t)$. The goal of the agent is to maximize the expected return $\mathbb{E}_\pi [\sum_{t=0}^\infty \gamma^t r_t]$, where $\gamma \in [0, 1)$ is the discount factor. 

In our framework, the lower-body policy is trained to achieve stable and robust locomotion, following conventional sim-to-real training pipelines for robust lower-body control works \cite{wang2025booster, gu2024humanoid, wu2025learn}. This allows it to handle diverse locomotion tasks without additional adaptation. \
The main difficulty lies in the upper–body control, which compensates for the disturbances induced by the freely moving base.
To make this problem tractable, we introduce two assumptions:

\noindent \textbf{Assumption 1: Negligible arm-to-base back–coupling.}  
    The arms are dynamically light relative to the lower body. Control actions in the arms induce negligible reaction forces on the base, allowing us to model base motion as an exogenous input when computing compensation torques.
    
\noindent \textbf{Assumption 2: Robust locomotion controller.} 
    The locomotion controller is robust enough to maintain balance and track desired trajectories despite disturbances generated by arm movements and control torques.

With the first assumption, we can simplify the disturbance model by treating the base motion as independent of arm movements. This allows us to focus solely on compensating for external base motion and ignore the lower-body's dynamic response to the upper-body. 
The second assumption allows us to design a controller without torque constraints. 
\subsection{Model-Enhanced Residual Learning}
We frame upper-body stabilization as controlling an arm subject to disturbances from a moving base.
Our method trains an RL residual policy that actively counteracts these locomotion-induced disturbances through a three-stage pipeline: 
(i) simulate base-induced inertial effects in a physically consistent manner, 
(ii) compute the analytic compensation torque that can cancel the resulting arm end-effector accelerations, and 
(iii) distill these signals into policy via reward reshaping, guiding it to output a joint command that stabilizes the arm end-effector.
The overall framework is illustrated in Fig.~\ref{fig:framework}.

\subsubsection{\textbf{Simulated Base Acceleration}}
\label{III-B} 
Directly applying spatial accelerations to a floating base in simulation is numerically unstable due to the nonlinear inverse dynamics problem.
Instead, we emulate base motion on a fixed-base model by injecting the \emph{equivalent fictitious wrench} that would be induced by a generic base twist $V_b = [v_b^\top; \omega_b^{\top} ]^\top \in se(3)$ and spatial acceleration $A_b = [\dot{v}_b^\top; \dot{\omega}_b^\top]^\top \in \mathbb{R}^6$.
Under Assumption 1, for each link of mass $m$ and inertia $I$ located at position $r$ relative to the base with local velocity $v$, the induced inertial force and torque are $F_b = m(-\dot{v}_b - \dot{\omega}_b \times r - \omega_b \times (\omega_b \times r) - 2\,\omega_b \times v)$ and $T_b = - I\,\dot{\omega}_b - \omega_b \times (I \omega_b)$.
The terms correspond respectively to linear, Euler, centrifugal, and Coriolis forces, plus angular-acceleration and gyroscopic torques. Together, they reproduce the accelerations experienced under real base motion, but can be applied stably to a fixed-base model.

To approximate locomotion-induced perturbations, we construct base acceleration signals from two characteristic sources:
(i) an impulse acceleration signal from the foot-ground reaction force, and (ii) a rhythmic sway from the body's center of mass (CoM) shifting with each step \cite{westervelt2003hybrid}. 
We represent the composed signal as
\vspace{-1mm}
\begin{equation}
\mathbf{A}_{b}(t) 
= \sum_{k=1}^{N} \left[ \underbrace{\boldsymbol{p}_{k}\; g(t; T_k)}_{\substack{\text{Foot contact} \\\text{impulse}}}
+  \underbrace{\boldsymbol{s}_{k} \; \sin\left(2\pi t/T_k + \phi_{k}\right)}_{\text{Periodic CoM swing}}\right],
\end{equation}
where $g(t; T_k)$ is a Gaussian impulse with standard deviation $\SI{0.01}{\s}$ and unit peak amplitude, $\boldsymbol{p}_k \in \mathbb{R}^6$ is the impulse amplitude, $\boldsymbol{s}_k \in \mathbb{R}^6$ is the oscillation amplitude, and $\phi_k$ is a phase offset. 
We sample disturbance parameters to ensure their coverage of a diverse range of realistic signals. 
The base motion periods $\{T_{k}\}_{k=1}^{K}$ are drawn from a log-uniform distribution in the range $[\SI{0.64}{\s},\SI{1.28}{\s}]$, covering the range of natural human-like gait cycles while avoiding bias toward short or long strides.
Impulse amplitudes $\boldsymbol{p}_k$ are sampled from $[\SI{-100}{\m/\s^2},\SI{100}{\m/\s^2}]^6$, spanning strong ground-contact transients.
Oscillation amplitudes $\boldsymbol{s}_k$ are sampled from $[\SI{-10}{\m/\s^2},\SI{10}{\m/\s^2}]^6 $, representing lateral and vertical CoM sways.
Phase offsets $\phi_k$ are sampled uniformly from $[-\pi,\pi]$ to generate diverse periodic acceleration profiles.

This random sampling procedure produces a rich set of disturbance profiles that capture the variability of realistic base acceleration signals. By repeatedly exposing the policy to such disturbances during training, we encourage it to learn compensation strategies that are robust to different walking styles, contact timings, and gait controllers.

\subsubsection{\textbf{Compensation Torque}}
In a fixed-base model, the local end–effector acceleration is
 $a_{ee}^{\text{loc}} = J(q) \ddot{q} + \dot{J}(q)\dot{q}$,
where $J(q) \in \mathbb{R}^{6 \times n}$ is the end-effector Jacobian at configuration $q \in \mathbb{R}^n$, and $\dot q \in \mathbb{R}^n, \ddot q \in \mathbb{R}^n$ are the joint velocity and acceleration with $n$-DoF of the arm.  
With base motion $(V_b,A_b)$, the global end–effector acceleration is
$a_{ee}^{\text{glob}} = a_{ee}^{\text{base}} + a_{ee}^{\text{loc}}$,
where 
$a_{ee}^{\text{base}} = \dot{v}_b + 2\,\omega_b \times v_{ee}^{\text{loc}} 
+ \omega_b \times (\omega_b \times r_{ee}^{\text{loc}}) 
+ \dot{\omega}_b \times r_{ee}^{\text{loc}}$ for given end-effector velocity $v_{ee}^{\text{loc}}$ and displacement  $r_{ee}^{\text{loc}}$ in the local frame. 

In addition, fictitious wrenches on the arm links (Sec.~\ref{III-B}) induce a local responsive acceleration
\begin{equation}
    a_{\text{resp}} = J(q) M^{-1}(q)\textstyle\sum_{i}^BJ_i(q)^\top [F^{i}_{b}\,^{\top}; T^i_b\,^{\top}]^{\top}
\end{equation}
where $B$ is the number of linkages, $M(q)$ is the joint–space inertia matrix, $J_i(q) \in \mathbb{R}^{6 \times n}$ is a Jacobian for link $i$ and $[F^{i}_{b}\,^{\top}; T^i_b\,^{\top}]^\top \in \mathbb{R}^6$ denotes the  fictitious wrench acting on a link $i$ from the base.

To cancel these effects under Assumption 2, we compute a compensating acceleration $a_{\text{comp}}$ by exerting a joint torque $\tau_{\text{comp}} \in \mathbb{R}^n$, such that
$a_{ee}^{\text{base}} + a_{\text{resp}} + a_{\text{comp}} \approx 0$.
Using the operational–space formulation, and the minimum-norm torque  \cite{siciliano2009robotics}\footnote{While we adopt the minimum–norm solution, alternative formulations are possible that can incorporate a constraint solver with additional cost functions and constraints.} renders
\begin{equation}
\tau_{\text{comp}} = -J(q)^\top \Lambda(q)\,\big(a_{ee}^{\text{base}} + a_{\text{resp}}\big),
\end{equation}
where $\Lambda(q) \in \mathbb{R}^{6 \times 6}$ is the operational–space inertia matrix.  
This torque is combined with task-oriented controller signal $\tau_{\text{task}}$  (e.g., operational space tracking \cite{khatib2003unified} of $x_{\text{des}}$ and $\dot x_{\text{des}}$) to stabilize the end-effector. 
 i.e. $\tau_{\text{task}} = J(q)^\top[\Lambda(q)\left(\bar{K}_p (x_{\text{des}} - x) + \bar{K}_d (\dot x_{\text{des}} - \dot x)\right) + Q(q,\dot q) + G(q)]$, where $Q(q,\dot q)$ is the operational-space centrifugal/Coriolis term, $G(q)$ is the operational-space gravitational term, and $\bar{K}_p, \bar{K}_d$ are operational-space gains.
\subsubsection{\textbf{Compensation Residual Policy Training}}
Directly deploying the analytically computed compensation torque $\tau_{\text{comp}}$ on hardware is infeasible, due to sensor noise, missing angular acceleration signals on the hardware IMU, and sim-to-real gap, such as motor delays and friction.
Instead, we train an RL policy which outputs joint targets to a low–level PD controller that matches the desired compensation torque $\tau_{\text{comp}}$. 
The observation space consists of the history of an end-effector command and proprioception data - IMU data ($\dot v_b$, $\omega_b$), upper-body joint states, and previous actions. 
Let $\tau_{\mathrm{PD}} \;=\; K_p\big(q_{\text{des}}-q\big)\;+\;K_d\big(\dot q_{\text{des}}-\dot q\big)$ be the torque generated by the PD loop from the policy's targets $q_{\text{des}}$, where target joint velocities are fixed to be zero $(\dot q_{\text{des}}=0)$. 
In addition, we add an operational-space tracking control torque $\tau_{\text{task}}$ to achieve ideal target-tracking behavior. 
To improve robustness, we inject observation noise and friction during training, and regularize both control effort and end-effector accelerations to discourage excessive actuation and jittering. 
The reward encourages the measured torque $\tau_{\text{measured}}$ to match the ideal torque $\tau_{\text{comp}} + \tau_{\text{task}}$: 
\begin{equation}\label{eq:tracking_reward}
\textstyle r_{\tau} = -\,\|{\tau}_{\text{measured}} - (\tau_{\text{comp}} + \tau_{\text{task}})\|_2,
\end{equation}
along with auxiliary rewards that penalize the global accelerations similar to \cite{li2025softa}, action smoothness, and tracking tolerances. The policy is trained with PPO~\cite{schulman2017proximal} using recurrent actor–critic networks with hidden sizes $[256, 128, 128]$.

In this work, we set an end-effector target in the local frame. 
In this case, the policy minimizes the local tracking error while stabilizing the end-effector under base perturbations. Although target tracking and stabilization objectives may conflict, we address this issue by adding a tolerance margin in the tracking reward functions described in Table.~\ref{tab:upper_reward}. This allows the policy to balance precise tracking with robust stabilization.

Finally, to achieve greater stability, the target can be specified in the world frame and converted into the local frame at runtime. However, this requires an accurate, real-time estimation of the robot's world pose, which we leave for future work.

\begin{table}[!t]
    \centering
    \scriptsize
    \begin{tabular}{lcc}
    \toprule
    \textbf{Components} &\textbf{Equations} & \textbf{Weights (stddev.)} \\
    \midrule
    Alive & 1 & 10 \\
    Position & $\exp(-\|r^{\text{cmd}} - r\|^2/\sigma_r^2)$* & $ 10 \;(0.1)$ \\
    Orientation & $\exp(-\|Q^{\text{cmd}} \ominus Q\|^2/\sigma_r^2)$* & $ 10 \; (0.1)$ \\  
    Torque guide & $\|\tau_{\text{measured}} - (\tau_{\text{comp}} + \tau_{\text{task}})\|$ & $-0.1$ \\
    \quad exp form &  {\scriptsize$\exp\left(-\|\tau_{\text{measured}} - (\tau_{\text{comp}} + \tau_{\text{task}})\|\right)$} & $5$ \\
    \midrule
    EE Lin. Acc. & $\|a_e\|$ & $ -0.1$ \\
    \quad exp form & $\exp(-\|a_e\|^2/\sigma_a^2)$ & $1.0\; (3.0)$ \\
    EE Ang. Acc. & $\|\alpha_e\|$ & $-0.01$ \\
    \quad exp form & $\exp(-\|\alpha_e\|^2/\sigma_\alpha^2)$ & $1.0\;  (10)$ \\
    \midrule
    Action rates & $\|a_{\text{prev}} - a_{\text{current}}\|$ & $-0.1$ \\
    \bottomrule
    \end{tabular}
    \caption{Summary of upper-body training rewards. *Note that for position and orientation tracking rewards, we assign a small tolerance of \SI{0.05}{\meter}  and \SI{0.1}{\radian} each. All norms in the table are $\text{L}_2$ norm. ($\ominus$: quaternion subtraction) 
    }
    \label{tab:upper_reward}
\end{table}

\subsection{Locomotion Training}
\label{III.C.loco-training}
Following state-of-the-art locomotion works \cite{radosavovic2024real,gu2024humanoid, wu2025learn}, the policy observation space consists of four components: (i) clock signals (sine/cosine of gait phase), (ii) proprioception (base angular velocity, joint states, previous actions), (iii) base velocity command, and (iv) 5-step observation history for short-term memory. The action space controls 13 lower-body joints via target positions tracked by PD controllers.

For the design of reward functions, we build upon the formulations provided in Booster Gym \cite{wang2025booster}, which include balance stability, smoothness, and velocity tracking task progress with carefully assigned weights. 
Robustness is improved by randomizing upper-body joint targets, end-effector mass, and environment parameters across episodes.
Locomotion policies are trained in IsaacLab using PPO. Both actor and critic are MLPs with hidden sizes $[256, 128, 128]$.

\section{EXPERIMENTS AND RESULTS}
In the experiments, we use the Booster T1 humanoid, which stands \SI{1.2}{\meter} tall and possesses $29$ degrees of freedom.
\vspace{-1mm}
\subsection{Simulation Results}
To systematically demonstrate the advantages of our framework, we address the following key questions:

\noindent \textbf{Q1}: Does our model-enhanced residual learning controller achieve superior end-effector stability?

\noindent \textbf{Q2}: How much does our perturbation generation strategy improve end-effector stability?

\noindent \textbf{Q3}: Does our perturbation generation strategy enable robust generalization to previously unseen locomotion controllers?

\begin{table}[t]
\setlength{\tabcolsep}{2pt}
\renewcommand\arraystretch{0.8}
\centering
\scriptsize 
\begin{tabular}{llcccc}
\toprule
\multicolumn{2}{c}{} 
  & \multicolumn{2}{c}{LinAcc (m/s\textsuperscript{2}) \textcolor{black}{{$\downarrow$}}}
  & \multicolumn{2}{c}{AngAcc (rad/s\textsuperscript{2}) \textcolor{black}{{$\downarrow$}}} \\
  \cmidrule(r){3-4} \cmidrule(r){5-6}
Task & Method 
  &  \phantom{0} mean  & \phantom{0} max \phantom{0} 
  & mean & max 
  \\
\midrule
\multirow{7}{*}{Stepping}
  & IK       & 5.73$\pm\textcolor{gray}{0.15}$ & 15.2$\pm\textcolor{gray}{0.29}$ & 18.1$\pm\textcolor{gray}{0.58}$ & 74.3$\pm\textcolor{gray}{1.44}$ 
  \\
  & RL w/o {\tiny Sim. Acc.}      & 4.01$\pm\textcolor{gray}{0.07}$ & 16.2$\pm\textcolor{gray}{0.70}$ & 18.5$\pm\textcolor{gray}{0.37}$ & 78.9$\pm\textcolor{gray}{3.27}$ 
  \\
  & RL w {\tiny Sim. Acc.}       & 3.91$\pm\textcolor{gray}{0.05}$ & 16.8$\pm\textcolor{gray}{1.14}$ & 18.8$\pm\textcolor{gray}{0.32}$ & 69.6$\pm\textcolor{gray}{3.90}$ 
  \\
  & Ours w/o $\tau_{\text{task}}$               & 3.35$\pm\textcolor{gray}{0.02}$ & 9.48$\pm\textcolor{gray}{0.34}$ & 16.4$\pm\textcolor{gray}{0.09}$ & 78.4$ \pm\textcolor{gray}{1.79}$ 
  \\
  & Ours w/o $r_\tau$               & 2.42$\pm\textcolor{gray}{0.04}$ & 6.40$\pm\textcolor{gray}{0.29}$ & 13.3$\pm\textcolor{gray}{0.38}$ & 69.3$\pm\textcolor{gray}{11.7}$ 
  \\
  & Ours w/o {\tiny Sim. Acc.}                     & 2.60$\pm\textcolor{gray}{0.04}$ & 8.41$\pm\textcolor{gray}{0.44}$ & 13.5$\pm\textcolor{gray}{0.30}$ & 57.8$\pm\textcolor{gray}{0.34}$ 
  \\
  & Ours                   & \textcolor{myred}{\textbf{2.26}}$\pm\textcolor{gray}{0.02}$ & \textcolor{myred}{\textbf{5.92}}$\pm\textcolor{gray}{0.20}$ & \textcolor{myred}{\textbf{11.9}}$\pm\textcolor{gray}{0.18}$ & \textcolor{myred}{\textbf{56.9}}$\pm\textcolor{gray}{0.46}$ 
  \\
\midrule

\multirow{7}{*}{Forward}
  & IK       & 5.28$\pm\textcolor{gray}{0.27}$ & 15.5$\pm\textcolor{gray}{0.63}$ & 17.3$\pm\textcolor{gray}{0.35}$ & 77.4$\pm\textcolor{gray}{4.78}$ 
  \\
  & RL w/o {\tiny Sim. Acc.}      & 3.54$\pm\textcolor{gray}{0.20}$ & 15.7$\pm\textcolor{gray}{1.57}$ & 16.7$\pm\textcolor{gray}{1.10}$ & 77.4$\pm\textcolor{gray}{7.42}$ 
  \\
  & RL w {\tiny Sim. Acc.}       & 3.41$\pm\textcolor{gray}{0.06}$ & 12.3$\pm\textcolor{gray}{0.85}$ & 16.3$\pm\textcolor{gray}{0.51}$ & 58.6$\pm\textcolor{gray}{3.95}$ 
  \\
  & Ours w/o $\tau_{\text{task}}$               & 2.67$\pm\textcolor{gray}{0.15}$ & 7.04$\pm\textcolor{gray}{0.58}$ & 12.0$\pm\textcolor{gray}{0.54}$ & 47.3$\pm\textcolor{gray}{1.85}$ 
  \\
  & Ours w/o $r_\tau$               & 2.38$\pm\textcolor{gray}{0.08}$ & 6.59$\pm\textcolor{gray}{1.64}$ & 13.1$\pm\textcolor{gray}{0.55}$ & 59.3$\pm\textcolor{gray}{7.70}$ 
  \\
  & Ours w/o {\tiny Sim. Acc.}                     & 2.45$\pm\textcolor{gray}{0.02}$ & 8.46$\pm\textcolor{gray}{0.33}$ & 12.3$\pm\textcolor{gray}{0.10}$ & \textcolor{myred}{\textbf{42.8}}$\pm\textcolor{gray}{0.14}$ 
  \\
  & Ours                   & \textcolor{myred}{\textbf{2.29}}$\pm\textcolor{gray}{0.01}$ & \textcolor{myred}{\textbf{5.56}}$\pm\textcolor{gray}{0.19}$ & \textcolor{myred}{\textbf{11.4}}$\pm\textcolor{gray}{0.11}$ & 43.5$\pm\textcolor{gray}{1.51}$ 
  \\
\midrule
\multirow{7}{*}{Lateral}
  & IK       & 5.95$ \pm\textcolor{gray}{0.12}$ & 18.6$\pm\textcolor{gray}{6.99}$ & 19.5$\pm\textcolor{gray}{6.62}$ & 94.0$\pm\textcolor{gray}{37.9}$ 
  \\
  & RL w/o {\tiny Sim. Acc.}      & 4.18$ \pm\textcolor{gray}{0.13}$ & 17.4$\pm\textcolor{gray}{0.59}$ & 19.2$\pm\textcolor{gray}{0.55}$ & 83.8$\pm\textcolor{gray}{1.75}$ 
  \\
  & RL w {\tiny Sim. Acc.}       & 5.74$\pm\textcolor{gray}{0.15}$ & 15.2$\pm\textcolor{gray}{0.29}$ & 18.1$\pm\textcolor{gray}{0.58}$ & 74.3$\pm\textcolor{gray}{1.44}$ 
  \\
  & Ours w/o $\tau_{\text{task}}$               & 3.43$\pm\textcolor{gray}{0.09}$ & 11.9$\pm\textcolor{gray}{0.24}$ & 17.4$\pm\textcolor{gray}{0.38}$ & 95.9$\pm\textcolor{gray}{1.76}$ 
  \\
  & Ours w/o $r_\tau$               & 2.67$\pm\textcolor{gray}{0.03}$ & 6.73$\pm\textcolor{gray}{0.30}$ & 13.4$\pm\textcolor{gray}{0.07}$ & 55.7$\pm\textcolor{gray}{1.66}$ 
  \\
  & Ours w/o {\tiny Sim. Acc.}                     & 3.21$\pm\textcolor{gray}{0.02}$ & 12.0$\pm\textcolor{gray}{0.21}$ & 14.4$\pm\textcolor{gray}{0.08}$ & 62.8$\pm\textcolor{gray}{1.96}$ 
  \\
  & Ours                    & \textcolor{myred}{\textbf{2.40}}$\pm\textcolor{gray}{0.00}$ & \textcolor{myred}{\textbf{6.03}}$\pm\textcolor{gray}{0.63}$ & \textcolor{myred}{\textbf{12.2}}$\pm\textcolor{gray}{0.06}$ & \textcolor{myred}{\textbf{54.8}}$\pm\textcolor{gray}{0.50}$ 
  \\

 \midrule
\multirow{7}{*}{Rotation}
  & IK       & 6.06$\pm\textcolor{gray}{0.50}$ & 16.7$ \pm\textcolor{gray}{2.17}$ & 21.0$\pm\textcolor{gray}{2.41}$ & 89.4$\pm\textcolor{gray}{8.40}$ 
  \\
  & RL w/o {\tiny Sim. Acc.}      & 4.87$\pm\textcolor{gray}{0.13}$ & 20.2$\pm\textcolor{gray}{2.76}$ & 22.7$\pm\textcolor{gray}{0.53}$ & 100.$\pm\textcolor{gray}{5.38}$ 
  \\
  & RL w {\tiny Sim. Acc.}       & 4.31$\pm\textcolor{gray}{0.09}$ & 17.8$\pm\textcolor{gray}{0.61}$ & 19.8$\pm\textcolor{gray}{0.28}$ & 72.5$\pm\textcolor{gray}{1.91}$ 
  \\
  & Ours w/o $\tau_{\text{task}}$               & 3.76$\pm\textcolor{gray}{0.56}$ & 8.90$\pm\textcolor{gray}{0.86}$ & 16.9$\pm\textcolor{gray}{1.15}$ & 67.4$\pm\textcolor{gray}{13.7}$ 
  \\
  & Ours w/o $r_\tau$               & 2.93$\pm\textcolor{gray}{0.01}$ & 7.48$\pm\textcolor{gray}{0.11}$ & 15.6$\pm\textcolor{gray}{0.08}$ & 72.8$\pm\textcolor{gray}{0.78}$ 
  \\
  & Ours w/o {\tiny Sim. Acc.}                     & 2.89$\pm\textcolor{gray}{0.03}$ & 9.67$ \pm\textcolor{gray}{0.75}$ & 14.5$\pm\textcolor{gray}{0.14}$ & 67.4$\pm\textcolor{gray}{1.29}$ 
  \\
  & Ours                  & \textcolor{myred}{\textbf{2.75}}$\pm\textcolor{gray}{0.01}$ & \textcolor{myred}{\textbf{7.13}}$\pm\textcolor{gray}{0.95}$ & \textcolor{myred}{\textbf{14.1}}$\pm\textcolor{gray}{0.09}$ & \textcolor{myred}{\textbf{66.6}}$\pm\textcolor{gray}{1.44}$ 
  \\
\bottomrule
\end{tabular}
\caption{Benchmark results (MuJoCo) on end-effector stability.}
\label{tab:sim_stability}
\end{table}

\begin{table*}[t]
\centering
\setlength{\tabcolsep}{4pt}
\renewcommand\arraystretch{0.4}
\begin{threeparttable}
\centering
\scriptsize 
\begin{tabular}{llcccccccc}
\toprule
&
& \multicolumn{4}{c}{With Trained Locomotion Policy}
& \multicolumn{4}{c}{With Unseen Locomotion Policy} \\
\cmidrule(r){3-6} \cmidrule(r){7-10}

\multicolumn{2}{c}{}
& \multicolumn{2}{c}{LinAcc (m/s\textsuperscript{2}) }
& \multicolumn{2}{c}{AngAcc (rad/s\textsuperscript{2}) }
& \multicolumn{2}{c}{LinAcc (m/s\textsuperscript{2}) }
& \multicolumn{2}{c}{AngAcc (rad/s\textsuperscript{2}) } \\
\cmidrule(r){3-4} \cmidrule(r){5-6} \cmidrule(r){7-8} \cmidrule(r){9-10}
Task & Method &
\phantom{0}mean & \phantom{0}max\phantom{0} &
mean & max &
\phantom{0}mean & \phantom{0}max\phantom{0} &
mean & max \\
\midrule
\multirow{7}{*}{Stepping}
  & RL (Pre-Train)       & 6.57$\pm\textcolor{gray}{0.27}$ & 18.2$\pm\textcolor{gray}{1.39}$ & 28.0$\pm\textcolor{gray}{0.34}$ & 84.4$\pm\textcolor{gray}{7.94}$ &  -- & -- & -- & --\\
  \\
  & RL (Co-Train)          & 5.81$\pm\textcolor{gray}{0.11}$ & 26.2$\pm\textcolor{gray}{1.07}$ & 24.8$\pm\textcolor{gray}{0.44}$ & 143.$\pm\textcolor{gray}{9.38}$
  & 10.6$\pm\textcolor{gray}{0.04}$ & 25.7$\pm\textcolor{gray}{0.46}$ & 48.5$\pm\textcolor{gray}{0.44}$ & 173.$\pm\textcolor{gray}{5.88}$ \\
  \\
  & Ours (w Pre-Train Loco. Policy)                    & 3.27$\pm\textcolor{gray}{0.28}$ & 9.89$\pm\textcolor{gray}{0.70}$ & 17.5$\pm\textcolor{gray}{0.69} $
  & 92.7$\pm\textcolor{gray}{17.1}$ & \multirow{3}{*}{5.32$\pm\textcolor{gray}{0.02}$} & \multirow{3}{*}{18.4$\pm\textcolor{gray}{0.69}$} & \multirow{3}{*}{26.1$\pm\textcolor{gray}{0.25}$ }& \multirow{3}{*}{94.0$\pm\textcolor{gray}{1.13}$}\\
  \\
  & Ours (w Co-Train Loco. Policy)                    & 3.07$\pm\textcolor{gray}{1.75}$ & 11.5$\pm\textcolor{gray}{0.42}$ & 20.4$\pm\textcolor{gray}{0.52}$ 
  & 158.$\pm\textcolor{gray}{1.17}$ & &  & & 
  \\
\midrule
\multirow{7}{*}{Forward} 
  & RL (Pre-Train)       & 5.30$\pm\textcolor{gray}{0.30}$ & 12.4$\pm\textcolor{gray}{0.29}$ & 21.3$\pm\textcolor{gray}{0.44}$ & 63.2 $\pm\textcolor{gray}{6.79}$ &
   -- & -- & -- & --\\
  \\
  & RL (Co-Train)            & 5.36$\pm\textcolor{gray}{0.06}$ & 22.5$\pm\textcolor{gray}{0.39}$ & 23.9$\pm\textcolor{gray}{0.24}$ & 142.$\pm\textcolor{gray}{0.87}$ 
  & 8.29$\pm\textcolor{gray}{0.05}$ & 23.7$\pm\textcolor{gray}{1.87}$ & 34.1$\pm\textcolor{gray}{0.13}$ & 177.$\pm\textcolor{gray}{11.56}$ \\
  \\
  & Ours (w Pre-Train Loco. Policy)    & 3.44$\pm\textcolor{gray}{0.14}$ & 11.5$\pm\textcolor{gray}{1.07}$ & 17.4$\pm\textcolor{gray}{1.86}$ & 90.1$\pm\textcolor{gray}{19.0}$
  &  \multirow{3}{*}{4.76$\pm\textcolor{gray}{0.06}$} &  \multirow{3}{*}{18.0$\pm\textcolor{gray}{1.76}$} &  \multirow{3}{*}{24.8$\pm\textcolor{gray}{0.27}$} &  \multirow{3}{*}{82.5$\pm\textcolor{gray}{0.68}$} \\
  \\
  & Ours (w Co-Train Loco. Policy)                    & 4.16$\pm\textcolor{gray}{0.05}$ & 11.6$\pm\textcolor{gray}{0.525}$ & 20.0$\pm\textcolor{gray}{0.34}$ 
  & 169.$\pm\textcolor{gray}{13.83}$ & &  & & 
  \\
\midrule
\multirow{7}{*}{Lateral} 
  & RL (Pre-Train)       & 6.50$\pm\textcolor{gray}{0.23}$ & 19.4$\pm\textcolor{gray}{0.35}$ & 28.2$\pm\textcolor{gray}{1.53}$ & 136.$\pm\textcolor{gray}{1.44}$ & -- & -- & -- & --\\
  \\
  & RL (Co-Train)            & 6.70$ \pm\textcolor{gray}{0.09}$ & 30.1$\pm\textcolor{gray}{0.45}$ & 27.7$\pm\textcolor{gray}{0.30}$ & 141.$\pm\textcolor{gray}{4.60}$ 
  & 9.28$\pm\textcolor{gray}{0.02}$ & 25.2$\pm\textcolor{gray}{0.29}$ & 40.2$\pm\textcolor{gray}{0.18}$ & 168.$\pm\textcolor{gray}{5.00}$ \\
  \\
  & Ours (w Pre-Train Loco. Policy)   & 3.74$\pm\textcolor{gray}{0.32}$ & 14.4$\pm\textcolor{gray}{2.73}$ & 17.0$\pm\textcolor{gray}{1.19}$ & 70.7$\pm\textcolor{gray}{6.83}$ 
  &  \multirow{3}{*}{4.97$\pm\textcolor{gray}{0.01}$} &  \multirow{3}{*}{18.1$\pm\textcolor{gray}{1.61}$} &  \multirow{3}{*}{25.2$\pm\textcolor{gray}{0.14}$} &  \multirow{3}{*}{88.0$\pm\textcolor{gray}{4.18}$} \\
  \\
  & Ours (w Co-Train Loco. Policy)                    & 4.05$\pm\textcolor{gray}{0.05}$ & 12.4$\pm\textcolor{gray}{0.49}$ & 21.9$\pm\textcolor{gray}{0.39}$ 
  & 156.$\pm\textcolor{gray}{12.8}$ & &  & & 
  \\
\midrule
\multirow{7}{*}{Rotation} 
  & RL (Pre-Train)       & 6.47$\pm\textcolor{gray}{0.47}$ & 23.2$\pm\textcolor{gray}{9.03}$ & 29.4$\pm\textcolor{gray}{0.16}$ & 154.$\pm\textcolor{gray}{7.44}$ &  -- & -- & -- & --\\
  \\
  & RL (Co-Train)            & 6.38$\pm\textcolor{gray}{0.14}$ & 24.0$\pm\textcolor{gray}{1.73}$ & 27.0$\pm\textcolor{gray}{0.26}$ & 135.$\pm\textcolor{gray}{15.5}$
  & 9.84$\pm\textcolor{gray}{0.04}$ & 27.8$\pm\textcolor{gray}{1.93}$ & 42.4$\pm\textcolor{gray}{0.40}$ & 187.$\pm\textcolor{gray}{8.38}$ \\
  \\
  & Ours (w Pre-Train Loco. Policy)    & 4.31$\pm\textcolor{gray}{0.10}$ & 17.8$\pm\textcolor{gray}{0.61}$ & 19.8$\pm\textcolor{gray}{0.28}$ & 72.5$\pm\textcolor{gray}{1.91}$ 
      &  \multirow{3}{*}{5.28$\pm\textcolor{gray}{0.12}$} &  \multirow{3}{*}{21.0$\pm\textcolor{gray}{1.82}$} &  \multirow{3}{*}{27.2$\pm\textcolor{gray}{0.29}$} &  \multirow{3}{*}{104.$\pm\textcolor{gray}{0.97}$ }
  \\
  \\
  & Ours (w Co-Train Loco. Policy)                    & 4.15$\pm\textcolor{gray}{0.02}$ & 12.32$\pm\textcolor{gray}{0.31}$ & 22.8$\pm\textcolor{gray}{0.10}$ 
  & 143.$\pm\textcolor{gray}{8.56}$ & &  & & 
  \\
\bottomrule
\end{tabular}
\rightline{\textbf{--}\,: The transfered upper-body policy has failed the locomotion policy due to excessive arm accleration.}
\caption{Benchmarking results (MuJoCo) on robustness.
\label{tab:sim_robust}
}
\end{threeparttable}
\end{table*}

To address the above questions, we evaluate our proposed framework using two key metrics: (1) \textbf{end-effector stability}, measuring the effectiveness of acceleration compensation; (2) \textbf{robustness}, reflecting the capability to adapt across diverse and unseen locomotion skills. 

For \textbf{end-effector stability}, we compare the acceleration compensation of several baselines: 
(1) an \textit{IK-based} approach
(2) \textit{learning-based} approaches \textit{with or without} our perturbation generation (denoted as RL w or w/o Sim. Acc.), trained in a fixed-base scenario,  and (3) our proposed SEEC framework \textit{without} operational space torque, and (4) our SEEC framework \textit{without} torque guide reward.

For \textbf{robustness}, we assess performance under different locomotion policies by comparing: (1) the \textit{Pre-Train} framework, trained with a specific pretrained locomotion policy provided beforehand, (2) the \textit{Co-Train} framework, where we adopt the training setup from \cite{li2025softa}, locomotion and manipulation policies are trained simultaneously, while having the same control frequency for fair comparison. For the evaluation, we replace the locomotion part of each trained framework with a new locomotion policy trained with Sec.~\ref{III.C.loco-training}, and compare the end-effector stability. We denote this experiment as testing ``With Unseen Locomotion Policy". 

To ensure a comprehensive evaluation, we design a diverse set of simulation tasks that expose the robot to distinct locomotion scenarios and dynamic variations, including: (a) stepping in place, (b) forward walking at \SI{0.4}{\meter\per\second}, (c) lateral walking at \SI{0.4}{\meter\per\second}, and (d) rotational walking at \SI{0.4}{\radian\per\second}. For each scenario, we perform three roll-outs and record the end-effector acceleration in the world frame. Additionally, for a fair comparison, all the methods share the same $K_p$ and $K_d$ gains for low-level PD control: $10.0$ and $0.5$, respectively.

\textit{Results and Analysis}:
As shown in Table \ref{tab:sim_stability}, our proposed SEEC framework outperforms the baselines in end-effector acceleration stability across most locomotion tasks. Note that removing either the operational space torque component or the torque-guided reward substantially degrades the performance. This shows that simulated base accelerations with these components lead to effective compensation learning. Additionally, among the three ablation components, removing the operational space torque leads to the largest performance degradation, 36.11\% for mean linear acceleration and 26.39\% for mean angular acceleration, likely because this term provides a precise tracking signal that enables the RL policy to focus on learning only the compensation term, thereby improving overall performance.

Table \ref{tab:sim_robust} further demonstrates superior robustness over both the pre-train and co-train baselines when evaluated under a previously unseen locomotion policy. The pre-training method fails in all cases due to excessive arm movements, as the hierarchical training paradigm restricts the state-space exploration for the manipulation policy. In addition, the co-trained method exhibits an average degradation of 57.45\% and 60.14\% for mean linear and angular acceleration under an unseen locomotion policy, whereas ours shows an average degradation of 34.40\% and 21.52\%. This may be because the co-training setup relies heavily on coordinated interaction between the upper-body and lower-body for acceleration compensation. 
\vspace{-1mm}
\subsection{Hardware results}
\subsubsection{End-effector acceleration comparison}
\begin{figure*}[htbp]
    \centerline{
    \subfigure[End-effector linear acceleration plots.]{
        \centering
        \includegraphics[width=0.49\linewidth]{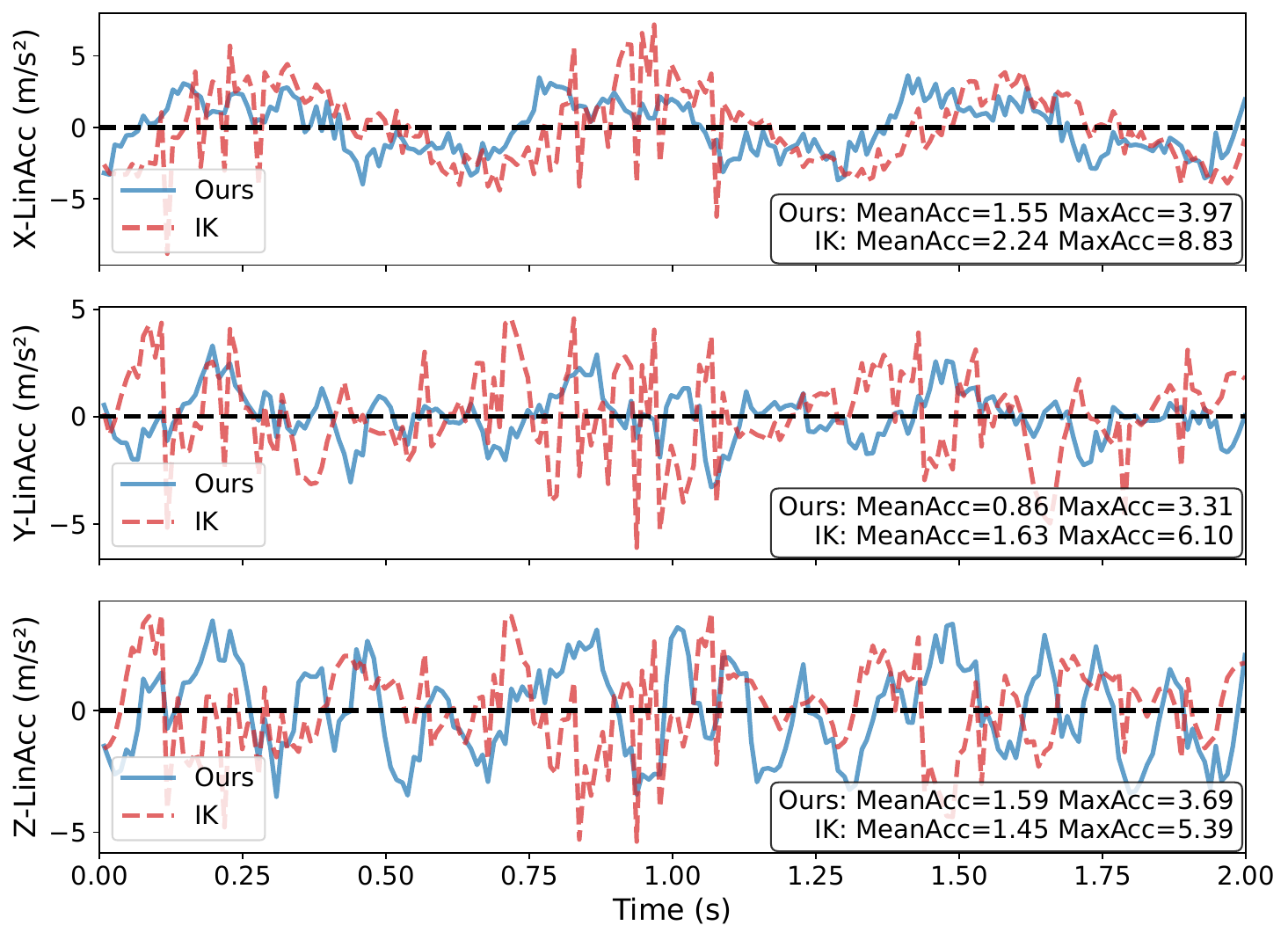}
        \label{fig:left}
        }
    \hfil
    \subfigure[End-effector angular acceleration plots.]{
        \centering
        \includegraphics[width=0.49\linewidth]{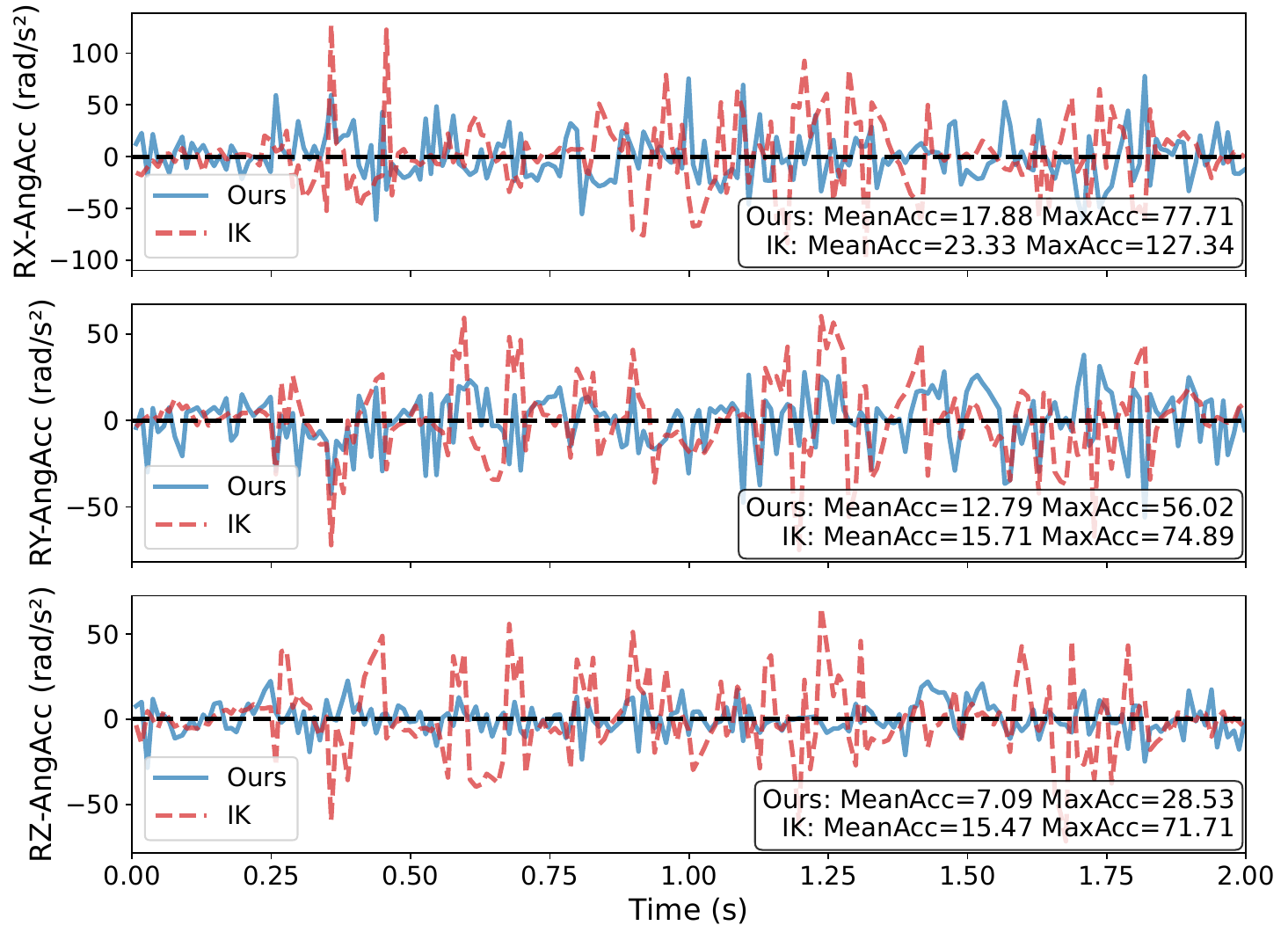}
        \label{fig:right}}
    }
    \caption{End-effector acceleration plots in real-world evaluation. The blue line indicates the acceleration profile of our method, and the dotted red line represents the baseline (IK) method.}
    \label{fig:acc_curve}
    \vspace{-2mm}
\end{figure*}

In hardware demonstrations, we deploy our SEEC framework on the T1 robot. To evaluate the effectiveness of our approach, we compare our controller against the IK-based baseline on the real robot and compute the end-effector acceleration from pose data collected by a motion capture system operating at \SI{120}{\Hz}, as in Table~\ref{tab:real_eval}. To obtain the end-effector acceleration, we apply numerical double differentiation to the recorded pose trajectories, removing abnormal or noisy measurements, as in Fig.~\ref{fig:acc_curve}. This provides a reliable measure of how the end-effector moves in a dynamically stable fashion.
\begin{table}[!ht]
\vspace{2mm}
\centering
\setlength{\tabcolsep}{3pt}
\renewcommand\arraystretch{0.6}
\scriptsize 
\begin{tabular}{lcccc}
\toprule
  & \multicolumn{2}{c}{LinAcc (m/s\textsuperscript{2}) \textcolor{black}{{$\downarrow$}}}
  & \multicolumn{2}{c}{AngAcc (rad/s\textsuperscript{2}) \textcolor{black}{{$\downarrow$}}} 
  \\
\cmidrule(r){2-3} \cmidrule(r){4-5}
Method 
  &  \phantom{0} mean  & \phantom{0} max \phantom{0} 
  & mean & max 
  \\
\midrule
   IK-Based Method      & 3.57 $ \pm\textcolor{gray}{0.46}$ & 11.6 $ \pm\textcolor{gray}{4.63}$ & 41.1 $ \pm\textcolor{gray}{4.31}$ & 151. $ \pm\textcolor{gray}{14.6}$ 
  \\
   SEEC (Ours)         & \textcolor{myred}{\textbf{2.82}} $ \pm\textcolor{gray}{0.11}$ & \textcolor{myred}{\textbf{6.36}} $ \pm\textcolor{gray}{0.36}$ & \textcolor{myred}{\textbf{24.2}} $ \pm\textcolor{gray}{4.62}$ & \textcolor{myred}{\textbf{78.6}} $ \pm\textcolor{gray}{9.81}$
  \\
\bottomrule
\end{tabular}
\caption{Real-world evaluation results on end-effector stability.}
\label{tab:real_eval}
\vspace{-5mm}
\end{table}
We observe consistent results as in the simulation, where both linear and angular accelerations remain stabilized over time. Notably, while the absolute acceleration magnitudes are in a similar range, our method shows a smoother acceleration profile, underscoring that our framework is more stable.

\subsubsection{Solving loco-manipulation tasks}
Furthermore, we test the proposed framework on tasks that require stable arm end-effector control under dynamic locomotion. The objective is to compensate for the hand acceleration when the robot is subject to whole-body motion and ground reaction forces. To this end, we design a set of representative tasks that combine locomotion with manipulation involving payloads.

\noindent \textbf{Chain Holding: } The T1 robot needs to grasp a chain with its hands and attempts to minimize its oscillation through stable motion control during walking. This task requires minimizing damping oscillations introduced by locomotion and highlights the robot's ability to regulate dynamic external objects, which is shown in Fig. \ref{fig:teaser}.

\noindent \textbf{Mobile Whiteboard Wiping:}  The robot needs to hold an eraser with its gripper and wipe a vertical whiteboard while continuously stepping, with VR teleoperation \cite{zhao2025xrobotoolkit}. The task requires the robot to maintain stable contact pressure and smooth wiping, ensuring effective cleaning while compensating for locomotion-induced disturbances.

\noindent \textbf{Plate Holding: } The T1  carries a plate of snacks while walking. This task requires minimizing plate acceleration to prevent spilling and demands precise stabilization of the upper-body during movement.

\noindent \textbf{Bottle Holding: } The robot carries a bottle of liquid while walking. To prevent spilling, the robot must suppress oscillations and avoid sudden accelerations.

\begin{figure}[h]
    \centering
    \includegraphics[width=\linewidth]{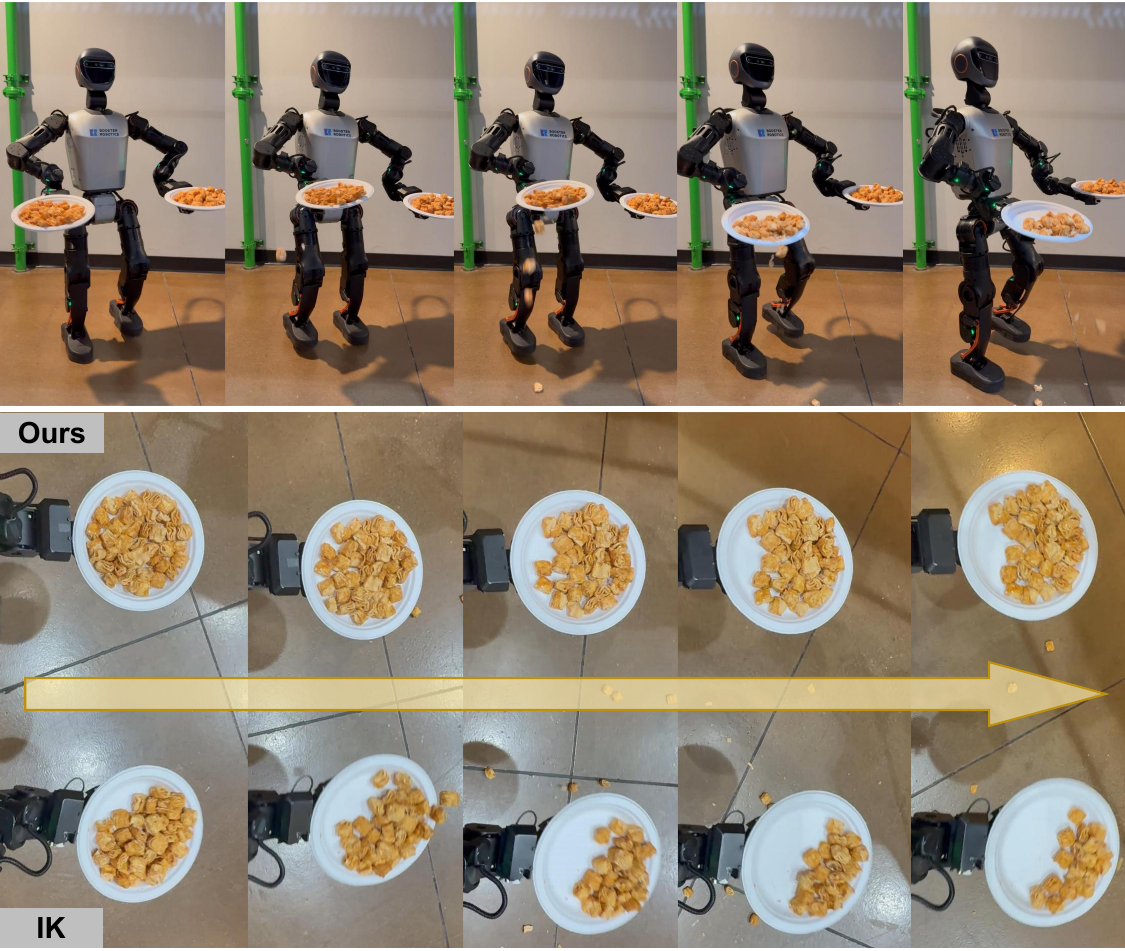}
    \caption{Plate holding task. With our method, the robot can stably hold the plate without spilling the snacks, whereas the IK-based method causes noticeable end-effector oscillations, leading to significant spillage.}
    \label{fig:plate_task}
    \vspace{-5mm}
\end{figure}

\begin{figure}[h]
    \centering
    \includegraphics[width=\linewidth]{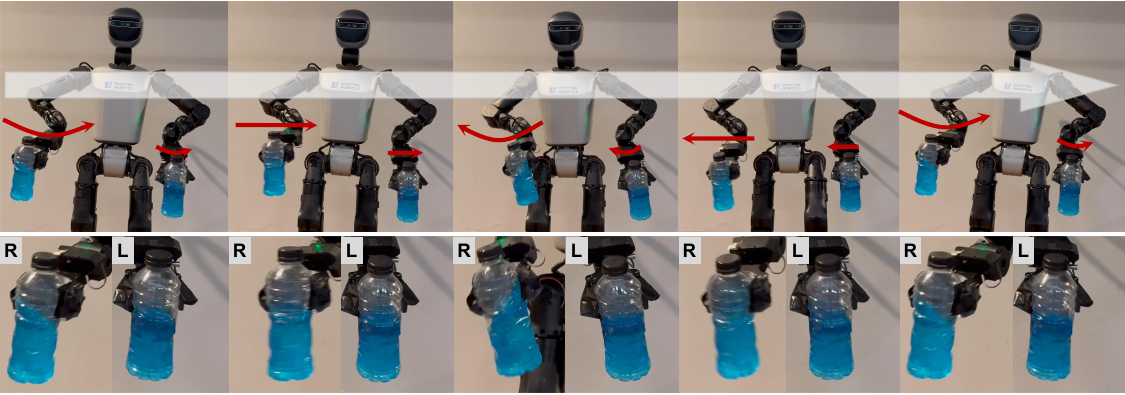}
    \caption{Bottle holding task. The left arm is controlled by our approach, achieving stable holding with minimal liquid surface vibration, while the right arm is controlled by the IK baseline, resulting in pronounced liquid oscillations.}
    \label{fig:bottle_task}
    \vspace{-5mm}
\end{figure}

\textit{Results and Analysis.}
For the \textbf{Chain Holding} task, as in Fig. \ref{fig:teaser} (A-D), without acceleration compensation, the chain exhibits large-amplitude oscillations due to base motion, leading to a final dropping off from the robot hand. With our SEEC framework, the robot effectively suppresses oscillatory dynamics, reducing oscillation amplitude and maintaining the chain nearly vertical during walking. This highlights the robustness of our framework in regulating external objects subject to dynamic excitations.
For the \textbf{Mobile Whiteboard Wiping} task in Fig. \ref{fig:teaser}(E), our framework consistently maintains smooth trajectories and steady end-effector contact forces, leading to clean wiping performance.
For the \textbf{Plate Holding} task, as shown in Fig.~\ref{fig:plate_task}, the baseline method spills the snacks as the robot walks. In contrast, our approach produces stable motions that allow the robot to carry the plate without spilling.
For the \textbf{Bottle Holding} task, as shown in Fig.~\ref{fig:bottle_task}, the time-lapse shows the liquid shaking violently, leading to a sudden splash, while our method keeps the bottle steady with minimized sloshing.

These real-world tasks verify that our framework enables stable end-effector control under dynamic locomotion. The results highlight its robustness to disturbances and dynamic loco-manipulation scenarios.

\section{CONCLUSIONS AND Discussion}

In this work, we introduce SEEC, a framework designed to achieve stable end-effector control for humanoid loco-manipulation. Our approach integrates model-based strategies into a learning-based end-effector controller, leveraging base acceleration data from simulation to enhance acceleration compensation. Experimental results from simulation demonstrate that our method consistently outperforms baseline approaches, yielding reduced end-effector acceleration and thereby improving stability.

Our method could benefit from the integration of more advanced model-based controllers and RL training strategies. While we have verified the effectiveness of model-enhanced residual learning, our model can benefit from a model-based controller that can handle constraints and would promote safe and stable operation, especially combined with constrained learning strategies.

Additionally, richer state inputs and more accurate state estimation would improve compensation and enable global target tracking. Our policy currently relies on upper-body proprioceptive input, which by design reacts to disturbances rather than proactively counteracting them. Whole-body state estimation could address this limitation and enable global target tracking to achieve more versatile loco-manipulation tasks, such as collaborative transports.


\section*{ACKNOWLEDGMENT}
 We would also like to express our gratitude for the hardware support provided by Booster Robotics.

\bibliographystyle{IEEEtran}
\bibliography{IEEEabrv,mybib}

\end{document}